\documentclass[10pt,twocolumn,a4paper]{esaAI}

\usepackage[table,xcdraw]{xcolor}
\usepackage{multirow}
\usepackage{booktabs}
\usepackage{eso-pic}

\newcommand{\setfoot}{
    \AddToShipoutPicture*{\AtPageLowerLeft{
        \put(0,40){\makebox[\paperwidth]{\hfill Accepted at SPAICE Conference, ECSAT, UK, 2024 \hfill}}
    }}
}

\title{CosmoCLIP: Generalizing Large Vision-Language Models for Astronomical Imaging}

\def\authorEmail{raza.imam@mbzuai.ac.ae}

\author[1]{Raza Imam\thanks{Corresponding author. E-Mail: \authorEmail}\thanks{Equal Contribution}}
\author[1]{Mohammed T. Alam$^{\dag}$}
\author[1]{Umaima Rahman}
\author[1]{Mohsen Guizani}
\author[1]{Fakhri Karray}
\affil[1]{Mohamed bin Zayed University of Artificial Intelligence, Abu Dhabi, UAE}

\AtBeginDocument{\setfoot}

\begin{document}

\makeCustomtitle

\begin{abstract}
Existing vision-text contrastive learning models enhance representation transferability and support zero-shot prediction by matching paired image and caption embeddings while pushing unrelated pairs apart. However, astronomical image-label datasets are significantly smaller compared to general image and label datasets available from the internet.
We introduce \texttt{CosmoCLIP}, an astronomical image-text contrastive learning framework precisely fine-tuned on the pre-trained CLIP \cite{radford2021learning} model using SpaceNet and BLIP-based \textit{descriptive} captions. SpaceNet, attained via FLARE \cite{alam2024flare}, constitutes $\sim$13k \textit{optimally} distributed images, while BLIP acts as a rich \textit{knowledge extractor}. The rich semantics derived from this SpaceNet and BLIP descriptions, when learned \textit{contrastively}, enable \texttt{CosmoCLIP} to achieve superior generalization across various \textit{in-domain} and \textit{out-of-domain} tasks. Our results demonstrate that \texttt{CosmoCLIP} is a straightforward yet powerful framework, significantly outperforming CLIP in zero-shot classification and image-text retrieval tasks.
\end{abstract}
\vspace{-0.75em}

\section{Introduction}
The introduction of new and advanced observational technologies in the field of astronomy has led to an unprecedented era of data proliferation. 
Large-scale sky surveys, such as the Sloan Digital Sky Survey (SDSS), the Dark Energy Survey (DES), and the upcoming Vera C. Rubin Observatory's Legacy Survey of Space and Time (LSST), are generating enormous volumes of high-resolution data, encompassing everything from galaxy surveys to deep-field observations \cite{ivezic2019lsst, dey2019overview}.
\vspace{-1.2em}
\paragraph{Motivation}
The vast amount of data being generated often lacks high-quality labels or representations, posing a significant challenge for astronomical analysis. Traditional data analysis methods, which rely heavily on manual inspection and interpretation, are becoming increasingly impractical. Consequently, a wide range of supervised and semi-supervised learning methods have been developed \cite{walmsley2022practical, zhang2022classifying}. However, these methods are constrained by their reliance on small training datasets, limiting their generalization to downstream tasks. This highlights the need for a large foundational model enriched with extensive multi-modal representations. These multi-modal representations when projected into a shared embedding space, can facilitate cross-modal relations, retrievals, and inferences, and can be effectively employed for various downstream tasks \cite{hackstein2023evaluation}.
\vspace{-1.2em}

\paragraph{Related Works}
Our study aims to address these challenges by integrating machine learning techniques, particularly in the realm of self-supervised and contrastive learning.
Recent contrastive learning methods like PAPERCLIP \cite{mishra2024paperclip} and AstroCLIP \cite{lanusse2023astroclip} have advanced astronomical analysis by linking image observations with natural language and aligning galaxy images with spectra, respectively. PAPERCLIP leverages abstracts from successful observing proposals and Hubble Space Telescope images, while AstroCLIP captures properties for red-shift and mass estimation. 
\cite{fu2024versatile} demonstrated a human-in-the-loop module with a large vision model for various tasks, including galaxy morphological classification, image restoration, object detection, and parameter extraction, showcasing few-shot learning capabilities \cite{slijepcevic2022learning,stein2021self,hayat2021self,chen2003improved,grill2020bootstrap, tanoglidis2021deepshadows, baek2024deep, sortino2023radio}.
\vspace{-1.2em}
\begin{figure*}[t]
    \centering
    \includegraphics[width=0.95\textwidth]{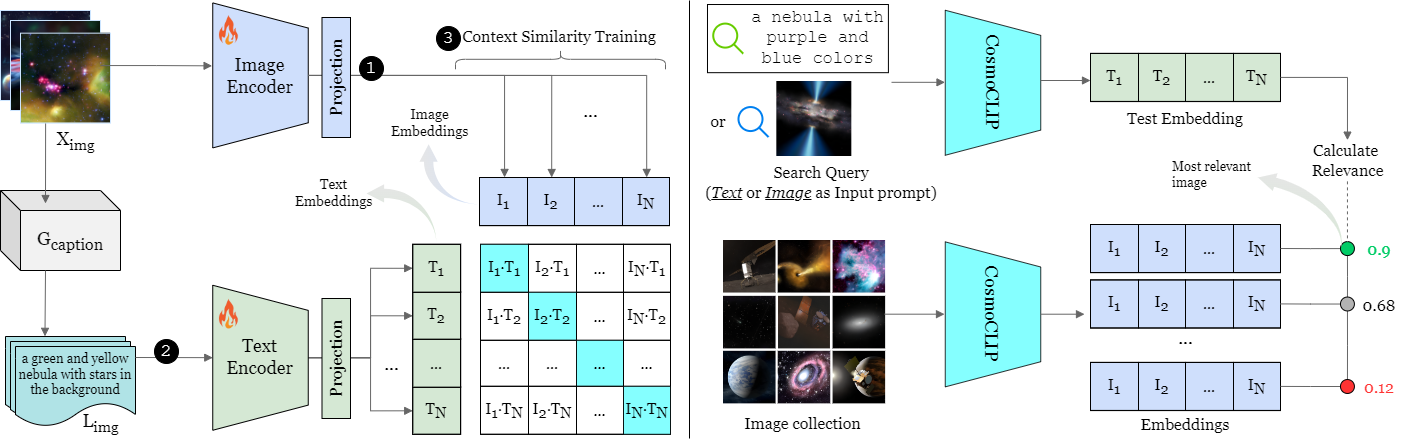}
    \caption{Overview of proposed framework: \texttt{CosmoCLIP}. (Left) For fine-tuning, inputs images from SpaceNet $X_{img}$ are processed via pre-trained CLIP image encoder to achieve image embeddings (1). BLIP $G_{caption}$ acts as a knowledge extractor, inputting SpaceNet images $X_{img}$ and generating descriptive captions $L_{img}$, which then act as image-text pairs for similarity training (3). (Right) Given an input text or image prompt, zero-shot prediction or image-text retrieval is performed.}
    \label{fig:method}
    \vspace{-0.40 cm}
\end{figure*}
\paragraph{Contribution}
To address aforementioned problem statement, (1) we propose to \textit{contrastively} train a large vision-language models (VLMs) like CLIP \cite{radford2021learning} for astronomical applications, using optimally distributed SpaceNet images \cite{alam2024flare} and descriptive captions generated by advanced language models (such as BLIP \cite{li2022blip} and LLAVA-7B \cite{liu2024visual}) as \textit{knowledge extractors}. These models provide detailed textual descriptions of astronomical images, enhancing the ability of the CLIP model using the rich semantics, to interpret and classify complex celestial phenomena.
(2) We analyzed that fine-tuning VLMs with the \textit{natural-language} descriptions from BLIP significantly improves its performance, demonstrating the critical role of high-quality textual data in training multi-modal models like CLIP. This approach could potentially revolutionize the way we analyze and interpret astronomical data.

\section{Method: CosmoCLIP}


This section provides technical details about \texttt{CosmoCLIP}, as shown in Figure \ref{fig:method}. \texttt{CosmoCLIP} includes three main components: (1) vision and text encoders, which extract embeddings; (2) knowledge extraction, which creates the context similarity matrix; and (3) context similarity training, which fine-tunes the pre-trained VLM.
\vspace{-0.15 cm}

\subsection{Vision-Text Encodings}
\texttt{CosmoCLIP} consists of two encoders designed to process text and visual inputs simultaneously. 
The visual encoder, denoted as $\mathcal{E}_{v}$, maps a visual input $X_{img}$ to a fixed-length embeddings $\boldsymbol{f}_v$. A projection head $\mathcal{V}$ then maps $\boldsymbol{f}_v$ to $\boldsymbol{F}_v \in \mathbb{R}^{n}$ as:
\vspace{-0.05 cm}
\begin{equation}
    \boldsymbol{f}_v = \mathcal{E}_{v}(X_{img}) \hspace{1cm} \Rightarrow  \hspace{1cm} \boldsymbol{F}_v = \mathcal{V}(\boldsymbol{f}_v)
    \label{eq:F_v}
\end{equation}
The text encoder, $\mathcal{E}_{t}$, processes text input, generating a latent textual embeddings $\boldsymbol{f}_t$ from the class labels $y$, followed by projection of raw embeddings $\boldsymbol{f}_t$ to $\boldsymbol{F}_t \in \mathbb{R}^{n}$ as:
\vspace{-0.30 cm}
\begin{equation}
    \boldsymbol{f}_t = \mathcal{E}_{t}(X_{img}) \hspace{1cm} \Rightarrow  \hspace{1cm} \boldsymbol{F}_t = \mathcal{T}(\boldsymbol{f}_t)
    \label{eq:F_t}
\end{equation}
where $\mathcal{T}$ is the projection head, while $\boldsymbol{F}_v$ and $\boldsymbol{F}_t$ are feature representations for images and texts.

\subsection{Knowledge Extraction}
Knowledge extraction in \texttt{CosmoCLIP} involves generating high-quality image-text pairs using a large-scale captioning model, such as BLIP, denoted as $G_{caption}$. Given an input image $X_{img}$, the captioner generates a caption $L_{img}$:
\vspace{-0.25 cm}
\begin{equation}
    L_{img} = G_{caption}(X_{img})
\end{equation}
By applying $G_{caption}$ to a large dataset, we obtain $N$ paired image-text samples:\\
\begin{equation}
    (X_{img}^i, L_{img}^i) = (X_{img}^i, G_{caption}(X_{img}^i)) ~ \text{$\forall$ $i$}=\{1,\ldots N\}
    \label{eq:caption}
\end{equation}
These pairs are used to construct a context similarity matrix, which is then utilized for context similarity training to fine-tune the pre-trained model.

\subsection{Context Similarity Training}
\label{sec:sim_training}
Context similarity training in \texttt{CosmoCLIP} aims to align image and text embeddings in a shared space, enhancing their mutual understanding. Following the Eq. \ref{eq:F_v} and \ref{eq:F_t}, the image and text feature representations ($\boldsymbol{F}_v$ and $\boldsymbol{F}_t$) are projected into a joint embedding space using learned projection matrices $W_v$ and $W_t$ to achieve normalized embeddings as:
\begin{equation}
    \Tilde{\boldsymbol{F}_v} = \frac{\boldsymbol{F}_v \cdot W_v}{\|\boldsymbol{F}_v \cdot W_v\|_2}, \quad \Tilde{\boldsymbol{F}_t} = \frac{\boldsymbol{F}_t \cdot W_t}{\|\boldsymbol{F}_t \cdot W_t\|_2}
\end{equation}
Next, the scaled pairwise cosine similarities between image and text embeddings are computed:
\begin{equation}
    \hat{y} = \exp(\tau) \cdot (\Tilde{\boldsymbol{F}_v} \cdot \Tilde{\boldsymbol{F}_t}^\top)
\end{equation}
The symmetric loss function is then calculated using cross-entropy (CE) loss \cite{mao2023crossentropy}, applied along both axes of the similarity matrix:
\begin{equation}
\mathcal{L} = \frac{1}{2} \left( \text{CE}(\hat{y}, y, \text{axis}=0) + \text{CE}(\hat{y}, y, \text{axis}=1) \right)
\end{equation}
For zero-shot inference, each text feature with class labels $y \in {1, 2, \cdots, M}$ is paired with the image feature. The prediction probability for class $y_i$ given input $X_{img}$ is expressed as:
\begin{equation}
p(y_i|{X_{img}}) = \frac{\hat{y}_i}{\sum_{j=1}^{C}\hat{y}_j}
\label{eq:inference}
\end{equation}
where $\hat{y}$ denotes the cosine similarity, and $\tau$ is the softmax temperature parameter.

{\renewcommand{\arraystretch}{1.2}
\begin{table*}[t]
\caption{Quantitative results for zero-shot classification presenting Top-1 accuracy of \texttt{CosmoCLIP} compared to baseline CLIP ViT-B/32 (\textit{bs.}). OOD Average indicates the \textit{out-of-domain} average results of four datasets.
}
\vspace{-0.75em}
\centering
\resizebox{\textwidth}{!}{%
\begin{tabular}{l|l|llll|l|l}
\hline
\rowcolor[HTML]{F2F2F2} 
\multicolumn{1}{c|}{\cellcolor[HTML]{F2F2F2}} & \multicolumn{1}{c|}{\cellcolor[HTML]{F2F2F2}\textbf{In-Domain}} & \multicolumn{5}{c|}{\cellcolor[HTML]{F2F2F2}\textbf{Out-Of-Domain (OOD)}} & \multicolumn{1}{c}{\cellcolor[HTML]{F2F2F2}} \\

\rowcolor[HTML]{F2F2F2} 
\multicolumn{1}{c|}{\multirow{-2}{*}{\cellcolor[HTML]{F2F2F2}\textbf{Method}}} & \multicolumn{1}{c|}{\cellcolor[HTML]{F2F2F2}\textbf{SpaceNet} \cite{alam2024flare}} & \multicolumn{1}{c}{\cellcolor[HTML]{F2F2F2}\textbf{Space} \cite{abhikalp_srivastava_space_images_category}} & \multicolumn{1}{c}{\cellcolor[HTML]{F2F2F2}\textbf{Spiral} \cite{altruistic_emphasis_spiral_galaxies}} & \multicolumn{1}{c}{\cellcolor[HTML]{F2F2F2}\textbf{Raw} \cite{alam2024flare}} & \multicolumn{1}{c}{\cellcolor[HTML]{F2F2F2}\textbf{Synthetic} \cite{alam2024flare}} & \multicolumn{1}{c|}{\cellcolor[HTML]{F2F2F2}\textbf{OOD Average}} & \multicolumn{1}{c}{\multirow{-2}{*}{\cellcolor[HTML]{F2F2F2}\textbf{Average}}} \\ 
\hline
CLIP (bs.) \cite{radford2021learning} & 6.45 & 8.07 & 5.65 & 8.36 & 4.45 & 6.63 & 6.60 \\
\texttt{CosmoCLIP}$_{LLaVA}$ & 27.97 & 21.66 & 90.08 & 30.59 & 26.08 & 42.10 & 39.28 \\
\rowcolor[HTML]{E2EFDA} 
\texttt{CosmoCLIP}$_{BLIP}$ & 70.87$~\uparrow$ & 62.12$~\uparrow$ & 95.68$~\uparrow$ & 45.72$~\uparrow$ & 83.36$~\uparrow$ & 71.72$~\uparrow$ & 71.55$~\uparrow$ \\ \hline
\end{tabular}}
\label{tab:zs_quantitative}
\vspace{-0.75em}
\end{table*}}

\section{Results and Discussion}
\subsection{Experimentation Details}
We conduct extensive experiments on one In-domain and four Out-of-Distribution datasets to answer the following questions:
\vspace{-0.75em}
\begin{enumerate}
    \item[Q1.] Does the knowledge-driven supervision, \textit{i.e.}, context similarity training yields better performance? \vspace{-0.75em}
    \item[Q2.] Does \texttt{CosmoCLIP} shows enhanced generalization performance for out-of-domain tasks with fine-tuning? \vspace{-0.75em}
    \item[Q3.] Are the learned embeddings efficient at Image-Text retrieval tasks? \vspace{-0.75em}
    \item[Q4.] How the learned embeddings of \texttt{CosmoCLIP} different from pre-trained CLIP? \vspace{-0.75em}
\end{enumerate}
\vspace{-1.0em}
\paragraph{Datasets} For fine-tuning the CLIP, we use our curated SpaceNet dataset \cite{alam2024flare}, which is \textit{in-domain} and contains about 12,900 samples across eight fine-grained classes: Planet, Asteroid, Nebula, Comet, Star, Black Hole, Galaxy, and Constellation. SpaceNet, representing an \textit{optimal} distribution, is created using our FLARE framework \cite{alam2024flare}. Leveraging traditional and diffusion-based augmentation, it helps address issues such as noisy backgrounds, low resolution, and distribution shifts, resulting in better generalization.
For testing \textit{out-of-domain} datasets, representing \textit{downstream tasks}, we use several existing datasets: Space \cite{abhikalp_srivastava_space_images_category}, Spiral \cite{altruistic_emphasis_spiral_galaxies}, Raw scraped samples from NASA \cite{nasawebsite}, and diffusion-based Synthetic images \cite{bao2023one}. These \textit{out-of-domain} datasets reflect real-world noisy distributions.
\vspace{-1.0em}

\paragraph{Implementation Details}
For fine-tuning, we initialized an ImageNet pre-trained CLIP-ViT-B/16 variant which includes a 12-layer Vision Transformer (ViT) for image encoding, producing 768-dimensional embeddings, and a 12-layer, 512-dimensional text encoder. This variant processes images at a resolution of 224$\times$224 pixels and handles text sequences up to 77 tokens. 
We implemented a linear projection head with 512 output dimensions and a learnable temperature parameter $\tau$ starting at 1.00. The other settings are: learning rate of 2e-5, batch size of 32, 10 epochs, and train:validation:test split of 80:10:10. We ran the entire setup on a single NVIDIA A100 40GB GPU. Additionally, following Eq. \ref{eq:caption}, we generated our textual descriptions via Image captioning models including LLaVa \cite{liu2024visual} and BLIP \cite{li2022blip}.

\subsection{Zero-Shot Classification}
We performed zero-shot image classification across various datasets, including the \textit{in-domain} SpaceNet and \textit{out-of-domain} distributions (Table \ref{tab:zs_quantitative}). Using Eq. \ref{eq:inference}, we employed trained image-text encoders to match test image embeddings with possible class prompt embeddings. Our \texttt{CosmoCLIP} model (\textit{i.e.}, \texttt{CosmoCLIP}$_{BLIP}$ trained on BLIP descriptions) consistently outperformed CLIP across all datasets. Notably, \texttt{CosmoCLIP} achieved a 64.42\% improvement over CLIP on the SpaceNet dataset (Q1), while across \textit{out-of-domain} tasks \texttt{CosmoCLIP} achieved a performance gain of 65.09\% than CLIP (Q2). Moreover, \texttt{CosmoCLIP}$_{LLaVA}$, trained on LLaVA descriptions, did not exhibit better generalization compared to \texttt{CosmoCLIP}$_{BLIP}$, underscoring the robustness of BLIP descriptions in training the model for better generalization.

\texttt{CosmoCLIP}, trained on SpaceNet data \cite{alam2024flare}, outperforms models trained on alternative datasets like scraped raw or synthetic data (Table \ref{tab:backbone_variants}). This superiority stems from optimal variance and favorable feature distribution in the feature space, enhancing training of LVMs such as CLIP. Additionally, Table \ref{tab:backbone_variants} demonstrates varying performance across different CLIP backbones, with SpaceNet-trained CLIP consistently showcasing superior generalization.

{\renewcommand{\arraystretch}{1.2}
\begin{table}[!t]
\caption{\texttt{CosmoCLIP} performance when fine-tuned on different datasets across different pre-trained backbones.}
\vspace{-0.75em}
\centering
\resizebox{0.5\textwidth}{!}{%
\begin{tabular}{l|lll|l}
\hline
\rowcolor[HTML]{F2F2F2} 
\textbf{CLIP Variants} & \textbf{Raw} & \textbf{Synthetic} & \textbf{SpaceNet} & \textbf{Average} \\ \hline
RN50 & 34.22 & 55.73 & 52.37 & 47.44 \\
ViT-B/32 & 37.25 & 70.64 & 83.00 & 63.63$~\uparrow$ \\
\rowcolor[HTML]{E2EFDA} 
ViT-B/16 & 40.87 & 73.26 & 84.50 & 66.21$~\uparrow$ \\
\hline
\end{tabular}}
\label{tab:backbone_variants}
\vspace{-0.75em}
\end{table}}

{\renewcommand{\arraystretch}{1.3}
\begin{table}[t]
\caption{Quantitative results for Image-Text Retrieval tasks presenting average cosine scores of top-\textit{k} similar embeddings.
}
\vspace{-0.75em}
\centering
\resizebox{0.5\textwidth}{!}{%
\begin{tabular}{c|l|llll} \hline
\multicolumn{1}{c|}{\cellcolor[HTML]{F2F2F2}\textbf{Retreival}} & \multicolumn{1}{c|}{\cellcolor[HTML]{F2F2F2}\textbf{Method}} & \cellcolor[HTML]{F2F2F2}\textbf{\textit{k}=1} & \cellcolor[HTML]{F2F2F2}\textbf{\textit{k}=3} & \cellcolor[HTML]{F2F2F2}\textbf{\textit{k}=5} & \cellcolor[HTML]{F2F2F2}\textbf{\textit{k}=10} \\ \hline
 & CLIP (bs.) & 54.02 & 53.90 & 53.80 & 53.55 \\
\multirow{-2}{*}{Img$\rightarrow$Img} & \cellcolor[HTML]{E2EFDA}\texttt{CosmoCLIP} & \cellcolor[HTML]{E2EFDA}93.60$~\uparrow$ & \cellcolor[HTML]{E2EFDA}92.34 & \cellcolor[HTML]{E2EFDA}91.53 & \cellcolor[HTML]{E2EFDA}90.46 \\ \hline
 & CLIP (bs.) & 21.09 & 20.83 & 20.64 & 20.41 \\
\multirow{-2}{*}{Txt$\rightarrow$Img} & \cellcolor[HTML]{E2EFDA}\texttt{CosmoCLIP} & \cellcolor[HTML]{E2EFDA}32.40$~\uparrow$ & \cellcolor[HTML]{E2EFDA}32.23 & \cellcolor[HTML]{E2EFDA}32.14 & \cellcolor[HTML]{E2EFDA}31.94 \\ \hline
\end{tabular}}
\label{tab:IT_retrieval}
\vspace{-0.75em}
\end{table}}

\subsection{Image-Text Retrieval}
Table \ref{tab:IT_retrieval} presents the results of our image-text retrieval experiments. \textit{Text-to-Image} retrieval involves inputting a user text query (such as $\langle$A realistic photo of a CLS$\rangle$) into \texttt{CosmoCLIP}'s text encoder, computing the cosine similarity between the encoded query and in-domain image samples, and retrieving the top-\textit{k} images with highest similarity score in order. Similarly, we perform it for an image input query for \textit{Image-to-Image} retrieval.

\begin{figure}[!t]
    \centering
    \begin{minipage}[b]{0.48\textwidth}
        \centering
        \includegraphics[width=\linewidth]{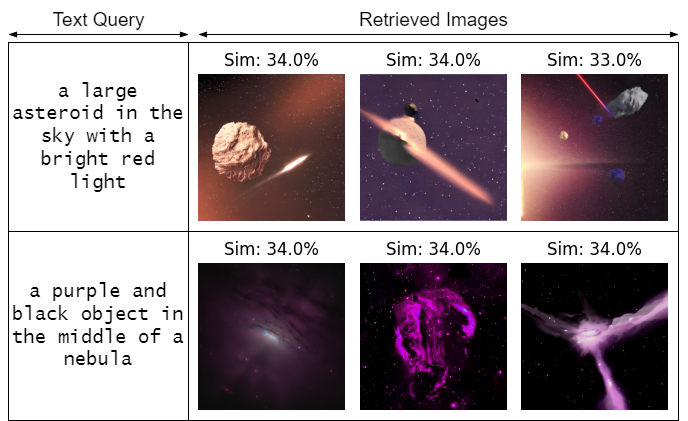}
        \caption*{(a) Text-to-Image Retrieval}
    \end{minipage}
    \hfill
    \begin{minipage}[b]{0.48\textwidth}
        \centering
        \includegraphics[width=\linewidth]{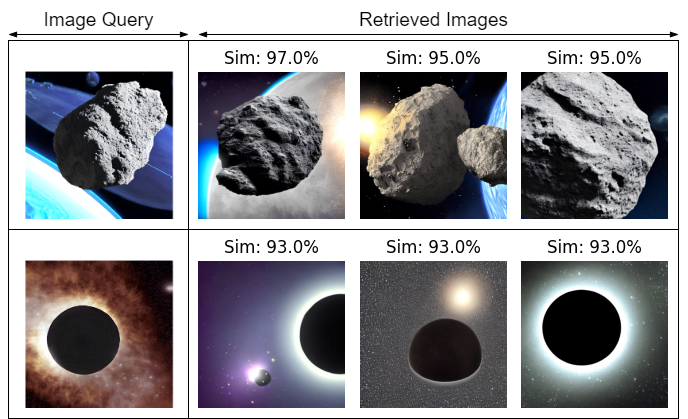}
        \caption*{(b) Image-to-Image Retrieval}
    \end{minipage}
    \caption{Qualitative results of \texttt{CosmoCLIP} for Text-to-Image retrieval (a) and Image-to-Image retrieval. \texttt{Sim} indicates Cosine Similarity between embeddings of input query and retrieved results.}
    \label{fig:retrieval_qualitative}
\vspace{-0.40em}
\end{figure}

\texttt{CosmoCLIP} consistently outperforms CLIP across all values of \textit{k}. For instance, at \textit{k}=1 for \textit{Text-to-Image} retrieval, \texttt{CosmoCLIP} achieves a score of 93.60 compared to 54.02 for CLIP. Similarly, for \textit{Text-to-Image} retrieval, where text prompts or queries are used to retrieve images, \texttt{CosmoCLIP} demonstrates superior performance over CLIP with +10 performance gain on average (Q3). These results highlight the effectiveness of \texttt{CosmoCLIP} in capturing meaningful semantic relationships between images and text, making it a promising model for various image-text retrieval tasks. Figure \ref{fig:retrieval_qualitative} shows qualitative Image-Text results of \texttt{CosmoCLIP}.

\begin{figure}[h]
    \centering
    \begin{minipage}[b]{0.23\textwidth}
        \centering
        \includegraphics[width=\linewidth]{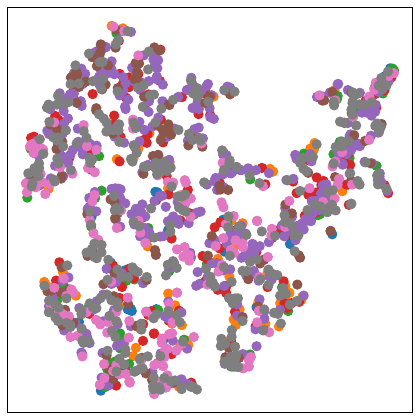}
        \caption*{(a) CLIP \cite{radford2021learning}}
    \end{minipage}
    \hfill
    \begin{minipage}[b]{0.23\textwidth}
        \centering
        \includegraphics[width=\linewidth]{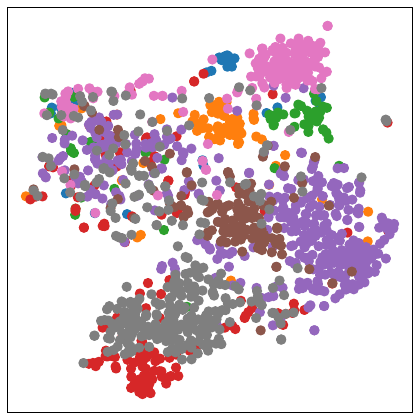}
        \caption*{(b) \texttt{CosmoCLIP} (Ours)}
    \end{minipage}
    \caption{t-SNE visualizations of the final class embedding for test set of SpaceNet data, representing 8 classes. \texttt{CosmoCLIP} could produce linearly separable features for generalization than pre-trained CLIP.}
    \label{fig:tsne_plots}
\vspace{-1.75em}
\end{figure}

\subsection{Representation Shift}
The t-SNE visualizations of CLIP and \texttt{CosmoCLIP}, extracted from visual token embeddings of the last layer of image encoder on the SpaceNet dataset, are depicted in Figure \ref{fig:tsne_plots}. Remarkably, \texttt{CosmoCLIP} showcases clearly discernible clusters, indicating a profound understanding of visual semantics. This robust separation suggests that \texttt{CosmoCLIP} has learned nuanced features that enable it to capture intricate patterns and nuances within the visual data, surpassing the capabilities of CLIP (Q4). Such enhanced representational capabilities signify \texttt{CosmoCLIP}'s potential to excel in diverse downstream tasks requiring nuanced understanding of visual content.

\section{Conclusion}
We present \texttt{CosmoCLIP}, an astronomical image-text contrastive learning framework, fine-tuned on the pre-trained CLIP model \cite{radford2021learning} using the optimally distributed SpaceNet dataset from the FLARE framework \cite{alam2024flare} and descriptive captions generated by BLIP captioning model. \texttt{CosmoCLIP} surpasses the baseline CLIP, achieving state-of-the-art performance across zero-shot classification and image-text retrieval tasks by a significant margin. With its rich feature semantics, \texttt{CosmoCLIP} is poised to become a foundational model for the astronomical domain, capable of handling a wide range of downstream tasks. In future, we plan to extend \texttt{CosmoCLIP}'s capabilities to encompass video analysis, further broadening its applicability and impact in the astronomical domain.

\newpage

\printbibliography
\addcontentsline{toc}{section}{References}

\end{document}